\pdfoutput=1

\documentclass[11pt]{article}

\usepackage[]{style/acl}

\usepackage{times}
\usepackage{latexsym}
\usepackage{multirow}
\usepackage[T1]{fontenc}

\usepackage{amsmath}
\usepackage{verbatim}
\usepackage{paralist}
\usepackage[utf8]{inputenc}

\usepackage{microtype}

\usepackage{inconsolata}

\usepackage{booktabs}
\usepackage{graphicx}
\usepackage[export]{adjustbox}
\usepackage{soul}

\usepackage{colortbl}
\usepackage{xcolor}
\usepackage{pgfplots}

\newcommand{\colorcell}[1]{
  \pgfmathparse{ifthenelse(#1 > 0, "green!" + #1 * 10 + "!white", "red!" + -#1 * 10 + "!white")}
  \edef\x{\noexpand\cellcolor{\pgfmathresult}}
  \x #1\%
}

\usepackage{tabularx}

\title{
  Paraphrasing in Affirmative Terms Improves Negation Understanding
}

\author{MohammadHossein Rezaei \and Eduardo Blanco \\
         Department of Computer Science, University of Arizona \\ \texttt{\{mhrezaei,eduardoblanco\}@arizona.edu}}

\begin{document}
\maketitle
\begin{abstract}
  Negation is a common linguistic phenomenon.
  Yet language models face challenges with negation in many natural language understanding tasks
  such as question answering and natural language inference.
  In this paper,
  we experiment with seamless strategies that incorporate affirmative interpretations (i.e., paraphrases without negation) to make models more robust against negation.
  Crucially, our affirmative interpretations are obtained automatically.
  We show improvements with CondaQA, a large corpus requiring reasoning with negation, and five natural language understanding tasks.
\end{abstract}

\section{Introduction}
\label{sec:introduction}

Negation is a fundamental linguistic phenomenon present in all human languages%
~\cite{Horn1989-HORANH}.
Language models underperform
in various natural language understanding (NLU) tasks when the input includes negation. 
For example, 
\citet{ettinger-2020-bert} and 
\citet{kassner-schutze-2020-negated} 
show that BERT~\citep{devlin-etal-2019-bert} 
fails to distinguish between negated and non-negated cloze questions.
Researchers have also shown that large language models such as GPT-3 
\cite{NEURIPS2020_1457c0d6}
and InstructGPT 
\cite{ouyang2022training}
are insensitive to negation
and fail to reason under negation 
\cite{truong-etal-2023-language}. 
\citet{jang-etal-2022-beyond}
point out that language models violate the logical negation property
($p$ is true iff $\neg p$ is false).
\citet{hossain-etal-2022-analysis}
analyze negation in
eight popular corpora for six NLU tasks.
They conclude that 
(a)~NLU corpora have few negations compared to general-purpose texts 
and
(b)~the few negations in them are often unimportant.
To our knowledge, CondaQA~\cite{ravichander-etal-2022-condaqa}
is the largest benchmark (14,182 question-answer pairs from Wikipedia)
requiring reasoning over the implications of negations.

In this paper, we paraphrase sentences with negation \emph{without using negation}
to make models for natural language understanding more robust when negation is present in the input.
We will use the term \emph{affirmative interpretation} to refer to paraphrases without negation
(e.g., \emph{I am not sad}: \emph{I am just ok}, \emph{I am happy}, etc.).
Appendix \ref{sec:affirmativeexamples} provides examples of how affirmative interpretations differ from simple paraphrases.

The main contributions of this paper are
(a)~strategies to generate and incorporate affirmative interpretations 
and
(b) experimental results demonstrating that doing so yields better results.\footnote{Code available at \url{https://github.com/mhrezaei1/paraphrase-affirmative} under Apache 2.0 license.}
In addition to CondaQA, we experiment with
five of the eight corpora analyzed by
\citet{hossain-etal-2022-analysis}:
CommonsenseQA~\cite{talmor-etal-2019-commonsenseqa}, 
STS-B~\cite{cer-etal-2017-semeval}, 
QNLI~\cite{rajpurkar-etal-2016-squad},
WiC~\cite{pilehvar-camacho-collados-2019-wic},
and 
WSC~\cite{levesque_winograd_2012}.\footnote{See examples from these corpora in Appendix \ref{sec:nlucorpora}.} 
We do not experiment with the other three corpora because
they do not contain any negation \cite[COPA]{roemmele-etal-2011-choice},
there is no difference in results when negation is present~\cite[QQP; 0.01 in macro F1]{cer-etal-2017-semeval},
or has already been shown~\cite{hossain-blanco-2022-leveraging} to benefit from affirmative interpretations~\cite[SST-2]{socher-etal-2013-recursive}.
The corpora we experiment with are in English.
\paragraph{Related Work}
\label{sec:relatedwork}

Early research on negation targeted detecting negating cues and
generating semantic representations,
usually by identifying the scope and focus
\cite{Morante2011NegationCues,morante-daelemans-2012-conandoyle,van-son-etal-2016-building,khandelwal-sawant-2020-negbert,truong-etal-2022-improving}.

More recent works bypass formal representations.
Instead, they make neural models robust when the input contains negation.
\citet{hosseini-etal-2021-understanding} combine unlikelihood training and syntactic data augmentation
to enhance the ability of BERT to understand negation with negated LAMA~\cite{kassner-schutze-2020-negated}.
\citet{singh-etal-2023-nlms} present a pretraining strategy designed for negation.
Unlike these works, we couple original inputs containing negation with affirmative interpretations.

The first work on affirmative interpretations was by
\citet{sarabi-etal-2019-corpus}.
\citet{hossain-etal-2022-question}
present AFIN, 
a corpus of $\approx$3,000 sentences with negations and their affirmative interpretations.
These two previous works are limited to generating affirmative interpretations from negations;
they do not provide extrinsic evaluations.
More recently, 
\citet{hossain-blanco-2022-leveraging}
present Large-AFIN, 
over 153,000 pairs of sentences with negation and their affirmative interpretations
obtained from parallel corpora via backtranslation.
In this paper, we present strategies to generate affirmative interpretations that do not require parallel corpora or a machine translation system.
Moreover, we demonstrate that incorporating affirmative interpretations yields better results with CondaQA
and five other natural language understanding tasks.

\section{Generating Affirmative Interpretations}
\label{sec:affirmative}
An affirmative interpretation generator is a system 
that takes a sentence with negation as its input 
and outputs an affirmative interpretation.
The task is similar to paraphrase generation with an additional constraint: the output must not contain negation.

We use two approaches to generate affirmative interpretations.
The first one is an off-the-shelf T5
\cite{t5paper}
fine-tuned by
\citet{hossain-blanco-2022-leveraging}
with Large-AFIN (Section \ref{sec:relatedwork}) to generate affirmative interpretations.
We refer to this model as \texttt{T5-HB}, 
and to the affirmative interpretations generated by \texttt{T5-HB} as \texttt{A\textsubscript{HB}}.

The second approach bypasses
the need for a large collection of pairs of sentences with negation and their affirmative interpretations.
It is based on the work by \citet{chatgpt_paraphraser},
who fine-tuned T5 on a paraphrase dataset
obtained with ChatGPT
(419,197 sentences and five paraphrases per sentence). 
We refer to this model as \texttt{T5-CG}.
Note that it is trained to generate paraphrases---not affirmative interpretations.
We obtain affirmative interpretations with \texttt{T5-CG} by
generating five paraphrases and selecting the first one that does not contain negation.
We refer to these affirmative interpretations as \texttt{A\textsubscript{CG}}.\footnote{At the time of writing, ChatGPT cannot reliably paraphrase without negation. See an example in Appendix \ref{sec:chatgpt}}
For examples of \texttt{A\textsubscript{HB}} and \texttt{A\textsubscript{CG}}, see Appendix~\ref{sec:afanalysis}. 

We use all negation cues in CondaQA to identify negation cues in our experiments.
CondaQA contains over 200 unique cues, including
single words (e.g., inaction, unassisted, unknown), 
affixal negations (e.g., dislike, unmyelinated, unconnected, inadequate, impartial),
and multiword expressions (e.g., a lack of, in the absence of, no longer, not at all, rather than).
They also include multiple part-of-speech tags
such as nouns (e.g., absence, nobody, inability),
adverbs (e.g., indirectly, involuntarily, unexpectedly),
determiners (e.g., neither, no, none),
and verbs (e.g., cannot, refuse, exclude).
\section{Experimental Results}

We use RoBERTa-Large \cite{liu2019roberta} as the base model.
In addition to experimenting with the original inputs for a task
(e.g., passage and question from CondaQA),
we couple the original input with one affirmative interpretation of the sentence with negation (if any; no change otherwise).
Affirmative interpretations are 
concatenated to the original input after the \texttt{<sep>} special token. 
Our approach is the same regardless of the type of negation. 
For implementation details,
see Appendix~\ref{sec:condaqatraining} and~\ref{sec:nlutraining}.

\begin{table*}[h]
	\small
    \centering
    \newcommand{\sig}{$^{\ast}$}
\setlength{\tabcolsep}{0.06in}
\small
\begin{tabular}{l r l l c@{\hspace{.15in}}  r@{}l c@{\hspace{0.10in}} r@{}l r@{}l r@{}l r@{}l}
\toprule
& & \multicolumn{2}{c}{Input Representation} && \multirow{2}{*}{Acc.} && & \multicolumn{8}{c}{Group Consistency} \\
\cmidrule{3-4} \cmidrule{9-16}
& \# Pars. & \multicolumn{1}{c}{Training} & \multicolumn{1}{c}{Testing} && &&& All && Par. && Sco. && Aff.& \\
\midrule
\multicolumn{14}{l}{From \citet{ravichander-etal-2022-condaqa}} \\
~~~~RoBERTa-Large &355M& P+Q & P+Q && 54.1 &&& 13.6 && 51.6 && 26.5 && 27.2 & \\
~~~~UnifiedQA-v2-Base &220M& P+Q & P+Q && 58.0 &&& 17.5 && 54.6 && 30.4 && 33.0 & \\
~~~~UnifiedQA-v2-Large &770M& P+Q & P+Q &&  66.7 &&& 30.2 && 64.0 && 43.7 && 46.5 &\\
~~~~UnifiedQA-v2-3B &3B& P+Q & P+Q && 73.3 &&& 42.2 && 72.8 && 55.7 && 57.2 &\\

\midrule
Our Implementation \\
~~~~RoBERTa-Large & 355M & P+Q & P+Q && 64.9 &&& 29.6 && 61.3 && 42.3 && 48.3 & \\ 
~~~~~~~~w/ sentence with neg. from P (S) && P+Q+S & P+Q+S && \cellcolor{green!2} 65.2 &&& \cellcolor{green!27} 31.1 && \cellcolor{red!26} 58.4 && \cellcolor{green!23} 44.1 && \cellcolor{green!10} 49.2 & \\
~~~~~~~~w/ 1st par. of S by \texttt{T5-CG}  (\texttt{S\textsubscript{CG}}) && P+Q+\texttt{S\textsubscript{CG}} & P+Q+\texttt{S\textsubscript{CG}} && \cellcolor{green!6} 65.7 &&& \cellcolor{red!22} 28.4 && \cellcolor{red!4} 60.8 && \cellcolor{green!1} 42.4 && \cellcolor{green!3} 48.6 & \\

\multicolumn{1}{l}{~~~~~~~~w/ Affirmative Interpretations} \\
&& P+Q+\texttt{A\textsubscript{HB}} & P+Q && \cellcolor{red!17} 62.8 &&& \cellcolor{red!61} 26.3 && \cellcolor{red!7} 60.5 && \cellcolor{red!40} 39.2 && \cellcolor{red!56} 43.3 & \\
&& P+Q+\texttt{A\textsubscript{HB}} & P+Q+\texttt{A\textsubscript{HB}} && \cellcolor{green!18} 67.1 & \sig && \cellcolor{green!33} 31.4 && \cellcolor{green!5} 61.9 &&\cellcolor{green!19} 43.8 && \cellcolor{green!27} 50.7 &  \\
&& P+Q+\texttt{A\textsubscript{CG}} & P+Q && \cellcolor{red!30} 61.3 &&& \cellcolor{red!80} 23.4 && \cellcolor{red!15} 59.6 && \cellcolor{red!58} \cellcolor{red!80} 37.8 && 37.8 & \\
&& P+Q+\texttt{A\textsubscript{CG}} & P+Q+\texttt{A\textsubscript{CG}} && \cellcolor{green!12} 66.4 & \sig && \cellcolor{green!39} 31.7 && \cellcolor{green!11} 62.6 && \cellcolor{green!29} 44.6 && \cellcolor{green!12} 49.4 & \\ 
&& P+Q+\texttt{A\textsubscript{HB}}+\texttt{A\textsubscript{CG}} & P+Q+\texttt{A\textsubscript{HB}}+\texttt{A\textsubscript{CG}} && \cellcolor{green!5} 65.6 &&& \cellcolor{green!9} 30.1 && \cellcolor{red!3} 60.9 && \cellcolor{green!18} 43.7 && \cellcolor{green!18} 49.9 & \\ \addlinespace
&& P+Q+\texttt{A\textsubscript{G}} & P+Q && \cellcolor{red!11} 63.6 &&& \cellcolor{red!53} 26.7 && \cellcolor{green!0} 61.4 && \cellcolor{red!45} 38.8 && \cellcolor{red!50} 43.9 & \\
&& P+Q+\texttt{A\textsubscript{G}} & P+Q+\texttt{A\textsubscript{HB}} && \cellcolor{red!4} 64.4 &&& \cellcolor{red!24} 28.3 && \cellcolor{red!36} 57.2 && \cellcolor{red!20} 40.7 && \cellcolor{red!23} 46.2 & \\
&& P+Q+\texttt{A\textsubscript{G}} & P+Q+\texttt{A\textsubscript{CG}} && \cellcolor{green!5} 65.6 &&& \cellcolor{green!13} 30.3 && \cellcolor{red!0} 61.3 && \cellcolor{green!1} 42.4 && \cellcolor{green!7} 49.0 & \\ \addlinespace
&& P+Q+\texttt{A\textsubscript{G}} or \texttt{A\textsubscript{HB}} & P+Q && \cellcolor{red!20} 62.5 &&& \cellcolor{red!72} 25.7 && \cellcolor{red!10} 60.1 && \cellcolor{red!48} 38.6 && \cellcolor{red!67} 42.4 & \\
&& P+Q+\texttt{A\textsubscript{G}} or \texttt{A\textsubscript{HB}} & P+Q+\texttt{A\textsubscript{HB}} && \cellcolor{green!6} 65.7 &&& \cellcolor{green!11} 30.2 && \cellcolor{red!1} 61.1 && \cellcolor{red!13} 41.3 && \cellcolor{green!6} 48.9 & \\
&& P+Q+\texttt{A\textsubscript{G}} or \texttt{A\textsubscript{CG}} & P+Q && \cellcolor{red!36} 60.6 &&& \cellcolor{red!80} 22.0 && \cellcolor{red!30} 57.9 && \cellcolor{red!80} 35.2 && \cellcolor{red!80} 36.8 &\\
&& P+Q+\texttt{A\textsubscript{G}} or \texttt{A\textsubscript{CG}} & P+Q+\texttt{A\textsubscript{CG}} && \cellcolor{green!15} 66.7 & \sig && \cellcolor{green!48} 32.2 && \cellcolor{green!8} 62.2 && \cellcolor{green!33} 44.9 && \cellcolor{green!29} 50.9 & \\
\bottomrule
\end{tabular}
    \caption{
        Results on the CondaQA test set.
        Q, P and S stand for question, passage and sentence with negation from P.
        \texttt{S\textsubscript{CG}} stands for the first paraphrase of S obtained with \texttt{T5-CG},
        without avoiding negations.
        An asterisk (`*') indicates statistically significant improvements (McNemar's test 
            \cite{mcnemar1947note},
            $p < 0.05$) with respect to not using affirmative interpretations 
	        (P+Q).
        UnifiedQA is fine-tuned with $\approx$1M~question-answer pairs from 20~corpora yet it does not outperform our best approach to incorporate affirmative interpretations (Accuracy: 66.7 vs. 67.1) unless it uses an order of magnitude more parameters (3B vs. 355M).
        The negated sentence (S) or a paraphrase that is not an affirmative interpretation (\texttt{S\textsubscript{CG}}) bring minor improvements compared to \texttt{A\textsubscript{HB}} and \texttt{A\textsubscript{CG}} affirmative interpretations.
    }
    \label{tab:condaresults}
\end{table*}

\subsection{CondaQA}
\label{sec:condaqaresults}

CondaQA 
\cite{ravichander-etal-2022-condaqa}
is a question-answering dataset that requires reasoning over negation.
It was created by 
asking crowdworkers to write questions about a negated sentence
within a paragraph retrieved from Wikipedia.
Crowdworkers also made three edits to the original paragraph:
\begin{compactenum}
    \item \textit{Paraphrase Edit}: Paraphrase the negation.
    \item \textit{Scope Edit}: Change the scope of the negation.
    \item \textit{Affirmative Edit}: Remove the negation.
\end{compactenum}
Additionally, 
they answered the question based on the original passage and all three edited passages.
(see examples in Appendix~\ref{sec:condaqaexample}).
Note that \emph{paraphrase} edits preserve meaning thus answers remain unchanged.
On the other hand, \emph{scope} edits change meaning but the answer may or may not remain the same.
Finally, \emph{affirmative} edits reverse meaning thus answers are also reversed.

\emph{Paraphrase} edits are not the same as our affirmative interpretations---crowdworkers
were not asked to paraphrase \emph{without using negation}.
We discovered, however, that 40.5\% of these edits satisfy our definition of affirmative interpretation.
We believe crowdworkers simply found it intuitive to paraphrase the negation without using negation.
We refer to these affirmative interpretations as \texttt{A\textsubscript{G}}~(\texttt{G}old)
and only use them for training purposes,
as using them at prediction time would be unrealistic.

Our evaluation reuses the metrics proposed by the authors of CondaQA: 
accuracy and group consistency.
Group consistency is the percentage of questions answered correctly
for all the passages in a group. 
The groups include the original passage and either all three or one of the edited passages.

Table~\ref{tab:condaresults} summarizes the experimental results
(see Appendix~\ref{sec:condaqadetailed} for additional results).
Our implementation of RoBERTa-Large obtains substantially better results than those by \citet[Acc.: 64.9 vs. 54.1]{ravichander-etal-2022-condaqa}.
Reviewing the training details revealed that the difference is that they stop training after ten epochs while we use early stopping and stop after 18 epochs. 

The best-performing model in terms of accuracy is
UnifiedQA-v2-3B
\cite{khashabi2022unifiedqav2},
which is a 3B-parameter T5 model pre-trained on 20 question-answering corpora data ($\approx$1M, Appendix \ref{sec:unifiedqacorpora}). 
Smaller versions of UnifiedQA (220M and 770M parameters) obtain substantially lower results despite being trained with the same corpora (Acc.: 58.0 and 66.7).
Our implementation of RoBERTa-Large using the question and passage as input almost rivals UnifiedQA-v2-Large (64.9 vs. 66.7)
despite the latter having twice the size and being fine-tuned with $\approx$1M question-answering pairs.

\begin{table*}
\small
\centering
    \begin{tabularx}{\textwidth}{p{.6in}XX} 
      \toprule
        & Negated sentence & Affirmative interpretation \\
      \midrule
      Adjective (48\%) & The island became \textit{completely uninhabited} by 1980 with the automation of the lighthouse. & The island became \textit{vacant} by the 1980s because of the automation of the lighthouse. \vspace*{6pt}\\ 
      & They are also made to work the company \textit{unpaid} as a form of "training". &	They are made to work the company \textit{free} as a form of "training". \\
      \midrule
      Verb (28\%) & Early Negro leagues were able to attract top talent but \textit{were unable} to retain them due to financial, logistical and contractual difficulties. &	Early Negro Leagues were able to attract top talent but \textit{failed} to retain them due to financial, logistical and contractual difficulties. \vspace*{6pt}\\
      & Although the original date is \textit{not used in modern times}, it has become an official holiday.	& Although the original date was \textit{used in the ancient times}, it has become an official holiday. \\
      \midrule
      Quantity (24\%) & But \textit{nobody outside of the Muslim world} made daily use of them before Stevin. &	\textit{Muslim groups were the only ones} to made daily use of them before Stevin. \vspace*{6pt} \\
      & However, he enjoyed it but \textit{not at that age}. & He enjoyed it at \textit{another age}. \\
      \midrule
      \multirow{2}{.6in}{Drop negation without further modifications (10\%)} & The \textit{unpopular} central government found itself in
      the difficult position of trying to gain support for spending cuts from the
      recalcitrant regional governments. &  The central government found itself in a
      difficult position trying to get support for spending cuts from recalcitrant
      regional governments. \vspace*{10pt}\\
      & Approximately 30\% of the acellular component of bone consists of organic matter, while roughly 70\% by mass is attributed to the \textit{inorganic} phase. &	Around 30\% of the acellular component of bone is made up by organic matter. \\
      \bottomrule
      \end{tabularx}
      \caption{
        \label{fig:af}
  Qualitative analysis of \texttt{A\textsubscript{HB}} affirmative interpretations that
  result in fixing errors made by the system not using affirmative interpretations with CondaQA 
  (P+Q vs. P+Q+\texttt{A\textsubscript{HB}}, Table \ref{tab:condaresults}).
  The affirmative interpretations rephrase in affirmative terms
  an adjective (48\%), a verb (28\%), or a quantity (24\%).
  We also observe that 10\% are erroneous as they simply drop the negated content.
      }
  \end{table*}

\begin{table}
    \centering
    \small
\begin{tabular}{l c@{\hspace{.10in}} r r }
	\toprule 
    && \multicolumn{1}{c}{{\% w/ negation}} & \multicolumn{1}{c}{{\% meaning-preserving}} \\
	\midrule
    \texttt{A\textsubscript{HB}} && $23$ & $64$ \\
    \texttt{A\textsubscript{CG}} && $46$ & $83$ \\
    \texttt{S\textsubscript{CG}} && $60$ & $90$ \\
	\bottomrule
\end{tabular}
    \caption{
        Qualitative analysis (100 samples from CondaQA) of affirmative interpretations
        (\texttt{A\textsubscript{HB}} and \texttt{A\textsubscript{CG}})
		and the first paraphrase by \texttt{T5-CG} without avoiding negation (\texttt{S\textsubscript{CG}}).
		Affirmative interpretations are less meaning-preserving,
		but the experimental results demonstrate that they are more beneficial (Table \ref{tab:condaresults}).
    }
    \label{tab:qualitativeanalysis}
\end{table}

Coupling the original input (passage and question)
with either
the sentence that contains negation (S) or
the first paraphrase obtained with \texttt{T5-CG} with no effort to avoid negation (\texttt{S\textsubscript{CG}}) brings minor improvements (64.9 vs. 65.2, 65.7).
More interestingly, incorporating affirmative interpretations brings statistically significantly better results
(64.9 vs. 67.1 (\texttt{A\textsubscript{HB}}), 66.4 (\texttt{A\textsubscript{CG}}) and 66.7 (\texttt{A\textsubscript{G}} or \texttt{A\textsubscript{CG}}/\texttt{A\textsubscript{CG}})).
We conclude the following from the results:

\begin{compactitem}
\item The benefits of affirmative interpretations are not due to pinpointing the sentence within the passage that is most relevant to answer the question (P+Q+S vs. P+Q+\texttt{S\textsubscript{CG}} vs. P+Q+\texttt{A\textsubscript{HB}}).

\item Training with affirmative interpretations is always beneficial as long as they are also used at prediction time.
  Note that we only use automatically obtained affirmative interpretations~(all but \texttt{A\textsubscript{G}}) at testing time.
  However, using both of them together does not yield better results
  (Acc.: 65.6 vs. 66.4 and 67.1).
\item At training time, complementing \texttt{A\textsubscript{G}} (available for $\approx$40\% of paraphrase edits) with \texttt{A\textsubscript{HB}} or \texttt{A\textsubscript{CG}} is beneficial (last and second-to-last block).
\end{compactitem}

\paragraph{Qualitative and Error Analysis}
\label{sec:qualitativeanalysis}

Manual analysis of 100 samples from CondaQA reveals that
\texttt{A\textsubscript{CG}} contains less negations than \texttt{A\textsubscript{HB}} (46\% vs. 23\%).
\texttt{A\textsubscript{CG}}, however, contains less meaning-preserving paraphrases (36\% vs. 17\%).
On the other hand, paraphrases in \texttt{S\textsubscript{CG}} rarely do not preserve meaning (10\%) but often include negation (60\%). (Table~\ref{tab:qualitativeanalysis}). 
Sometimes it is not natural to rewrite a sentence without negation (e.g., \emph{The inner membrane is rich in an unusual phospholipid, cardiolipin.})
Out of the 23 samples where \texttt{A\textsubscript{HB}} contains negation, a human was able to rewrite 15 of them without negation.
Combined with the results from Table \ref{tab:condaresults},
this analysis leads to the conclusion that affirmative interpretations are beneficial despite being noisy.

We also analyzed 50 samples of the errors made representing the input with P+Q that are fixed using affirmative interpretations from \texttt{A\textsubscript{HB}}.
A negated adjective is replaced by its affirmative counterpart (e.g., \emph{not happy} $\rightarrow$ \emph{sad}) in 48\% of cases.
Table~\ref{fig:af} shows the analysis and examples of negated sentences and their \texttt{A\textsubscript{HB}}
affirmative interpretations.
\subsection{Other NLU Tasks}
\label{sec:nluresults}

We experiment with five additional NLU tasks to evaluate 
the benefits of affirmative interpretations. 
We access these corpora through the
GLUE
\cite{wang-etal-2018-glue}
and SuperGLUE
\cite{superglue}
benchmarks.
We report results on the development set of each corpus,
given that the test sets are not publicly available.
In addition,
we report the results for important and non-important instances 
as identified by \citet{hossain-etal-2022-analysis}.
They consider a negation \emph{unimportant} if one can disregard it and still make the correct prediction.
For example, \emph{John didn't eat the steak with gusto} (most likely) entails \emph{John ate meat} even if one disregards the negation.

\begin{table*}
    \centering
    \setlength{\tabcolsep}{0.045in}
\small
\begin{tabular}{l @{}c@{\hspace{.1in}} @{}l@{} r@{}l@{} r@{}l r@{}l r@{}l r@{}l@{} r@{}l r@{}l@{} r@{}l r@{}l@{} r@{}l}
\toprule
& & & \multicolumn{4}{c}{CmnsnsQA} & \multicolumn{4}{c}{STS-B} & \multicolumn{4}{c}{QNLI} & \multicolumn{4}{c}{WiC} & \multicolumn{4}{c}{WSC} \\
\cmidrule(lr){2-7} \cmidrule(lr){8-11} \cmidrule(lr){12-15} \cmidrule(lr){16-19} \cmidrule(lr){20-23} & & & F1 & & & & Prsn & & Sprmn & & F1 & & & & F1 & & & & F1 & & & \\
\midrule RoBERTa & & & 0.70 & & & & 0.92 & & 0.92 & & 0.93 & & & & 0.71 & & & & 0.69 & & & \\
~~~~instances without negation & & & 0.69 & & & & 0.92 & & 0.92 & & 0.93 & & & & 0.71 & & & & 0.67 & & & \\
~~~~ instances with negation & & & 0.73 & & & & 0.88 & & 0.88 & & 0.92 & & & & 0.66 & & & & 0.71 & & & \\
~~~~~~~~Important & & & 0.67 & & & & 0.82 & & 0.85 & & 0.78 & & & & n/a & & & & n/a & & & \\
~~~~~~~~Unimportant & & & 0.80 & & & & 0.88 & & 0.88 & & 0.92 & & & & 0.66 & & & & 0.71 & & & \\
\midrule \multicolumn{1}{l}{RoBERTa w/ Affirmative Interpret.} \\
~~~~obtained using \texttt{T5-HB} (\texttt{A\textsubscript{HB}}) & & &\multicolumn{2}{r}{\cellcolor{green!21}  0.72 }&\cellcolor{green!21} \ (+2.9\%) & & 0.92 & & 0.91 & &\multicolumn{2}{r}{\cellcolor{green!8}  0.94 }&\cellcolor{green!8} \ (+1.1\%) & &\multicolumn{2}{r}{\cellcolor{red!10}  0.70 }&\cellcolor{red!10} \ (-1.4\%) & &\multicolumn{2}{r}{\cellcolor{red!10}  0.68 }&\cellcolor{red!10} \ (-1.4\%) & \\
~~~~~~~~ instances without negation & & &\multicolumn{2}{r}{\cellcolor{green!32}  0.72 }&\cellcolor{green!32} \ (+4.3\%) & & 0.92 & & 0.92 & &\multicolumn{2}{r}{\cellcolor{green!8}  0.94 }&\cellcolor{green!8} \ (+1.1\%) & &\multicolumn{2}{r}{\cellcolor{red!0}  0.71 }&\cellcolor{red!0} \ (+0.0\%) & &\multicolumn{2}{r}{\cellcolor{red!56}  0.62 }&\cellcolor{red!56} \ (-7.5\%) & \\
~~~~~~~~ instances with negation & & &\multicolumn{2}{r}{\cellcolor{green!10}  0.74 }&\cellcolor{green!10} \ (+1.4\%) & & 0.88 & & 0.88 & &\multicolumn{2}{r}{\cellcolor{red!0}  0.92 }&\cellcolor{red!0} \ (+0.0\%) & &\multicolumn{2}{r}{\cellcolor{green!45}  0.70 }&\cellcolor{green!45} \ (+6.1\%) & &\multicolumn{2}{r}{\cellcolor{green!31}  0.74 }&\cellcolor{green!31} \ (+4.2\%) & \\
~~~~~~~~~~~~Important & & &\multicolumn{2}{r}{\cellcolor{green!33}  0.70 }&\cellcolor{green!33} \ (+4.5\%) & & 0.83 & & 0.84 & &\multicolumn{2}{r}{\cellcolor{green!80}  0.89 }&\cellcolor{green!80} \ (+14.1\%) & & n/a & &\ & & n/a & &\ & \\
~~~~~~~~~~~~Unimportant & & &\multicolumn{2}{r}{\cellcolor{red!0}  0.80 }&\cellcolor{red!0} \ (+0.0\%) & & 0.87 & & 0.88 & &\multicolumn{2}{r}{\cellcolor{red!0}  0.92 }&\cellcolor{red!0} \ (+0.0\%) & &\multicolumn{2}{r}{\cellcolor{green!45}  0.70 }&\cellcolor{green!45} \ (+6.1\%) & &\multicolumn{2}{r}{\cellcolor{green!31}  0.74 }&\cellcolor{green!31} \ (+4.2\%) & \\
~~~~obtained using \texttt{T5-CG} (\texttt{A\textsubscript{CG}}) & & &\multicolumn{2}{r}{\cellcolor{green!10}  0.71 }&\cellcolor{green!10} \ (+1.4\%) & & 0.92 & & 0.92 & &\multicolumn{2}{r}{\cellcolor{green!8}  0.94 }&\cellcolor{green!8} \ (+1.1\%) & &\multicolumn{2}{r}{\cellcolor{green!21}  0.73 }&\cellcolor{green!21} \ (+2.8\%) & &\multicolumn{2}{r}{\cellcolor{green!21}  0.71 }&\cellcolor{green!21} \ (+2.9\%) & \\
~~~~~~~~instances without negation & & &\multicolumn{2}{r}{\cellcolor{green!21}  0.71 }&\cellcolor{green!21} \ (+2.9\%) & & 0.93 & & 0.92 & &\multicolumn{2}{r}{\cellcolor{green!8}  0.94 }&\cellcolor{green!8} \ (+1.1\%) & &\multicolumn{2}{r}{\cellcolor{green!21}  0.73 }&\cellcolor{green!21} \ (+2.8\%) & &\multicolumn{2}{r}{\cellcolor{green!11}  0.68 }&\cellcolor{green!11} \ (+1.5\%) & \\
~~~~~~~~instances with negation & & &\multicolumn{2}{r}{\cellcolor{green!10}  0.74 }&\cellcolor{green!10} \ (+1.4\%) & & 0.88 & & 0.88 & &\multicolumn{2}{r}{\cellcolor{red!0}  0.92 }&\cellcolor{red!0} \ (+0.0\%) & &\multicolumn{2}{r}{\cellcolor{green!45}  0.70 }&\cellcolor{green!45} \ (+6.1\%) & &\multicolumn{2}{r}{\cellcolor{green!42}  0.75 }&\cellcolor{green!42} \ (+5.6\%) & \\
~~~~~~~~~~~~Important & & &\multicolumn{2}{r}{\cellcolor{green!22}  0.69 }&\cellcolor{green!22} \ (+3.0\%) & & 0.82 & & 0.87 & &\multicolumn{2}{r}{\cellcolor{green!80}  0.89 }&\cellcolor{green!80} \ (+14.1\%) & & n/a & &\ & & n/a & &\ & \\
~~~~~~~~~~~~Unimportant & & &\multicolumn{2}{r}{\cellcolor{red!0}  0.80 }&\cellcolor{red!0} \ (+0.0\%) & & 0.88 & & 0.88 & &\multicolumn{2}{r}{\cellcolor{red!0}  0.92 }&\cellcolor{red!0} \ (+0.0\%) & &\multicolumn{2}{r}{\cellcolor{green!45}  0.70 }&\cellcolor{green!45} \ (+6.1\%) & &\multicolumn{2}{r}{\cellcolor{green!42}  0.75 }&\cellcolor{green!42} \ (+5.6\%) & \\
\bottomrule
\end{tabular}

    \caption{
        Results on additional NLU tasks (macro F1 except with STS-B (Pearson and Spearman correlations)).
        Percentages between parentheses indicate improvements compared to models not using affirmative interpretations.
        Affirmative interpretations yield better results, and \texttt{A\textsubscript{CG}} outperforms \texttt{A\textsubscript{HB}}.
        The largest gains are with important negations, although we observe gains with instances without negation (up to 4.3\%) except with WSC (-7.5\%).
    }
    \label{tab:nluresults}
\end{table*}

Table~\ref{tab:nluresults} presents the results.
Incorporating affirmative interpretations
(\texttt{A\textsubscript{HB}} or \texttt{A\textsubscript{CG}})
improves performance across all corpora with instances containing important negations;
the only exception is STS-B with \texttt{A\textsubscript{HB}} (Spearman: -1.2\%)
and \texttt{A\textsubscript{CG}} (Pearson: no difference).
It is worth noting that WiC and WSC have no important negations,
yet either \texttt{A\textsubscript{HB}} or \texttt{A\textsubscript{CG}}
yield substantial improvements with unimportant negations (4.2--6.1\%). 
Surprisingly, we found that incorporating affirmative interpretations is beneficial for instances \emph{without} negation across all corpora except WSC with \texttt{A\textsubscript{HB}}.

These experiments demonstrate that incorporating affirmative interpretations
not only obtains higher or comparable results with instances containing important negations,
but also often improves results with instances not containing negation.
\section{Conclusion}
\label{sec:conclusion}
We have presented two strategies to generate and incorporate affirmative interpretations into models for natural language understanding.
The idea is simple yet effective: complement inputs that contain negation with a paraphrase that does not contain negation.
Crucially, we have demonstrated that automatically obtained (noisy) affirmative interpretations yield improvements with
(a)~CondaQA compared with a model with twice as many parameters pre-trained with $\approx$1M question-answer pairs from 20 existing corpora
and
(b) five NLU tasks.
Our methodology is architecture- and task-agnostic.
In fact, the model to generate affirmative interpretations was tuned with out-of-domain corpora.

\paragraph{Future Work.} 
The methods we have presented are simple and effective, but they are not the only way to incorporate or generate affirmative interpretations.
For example, one might be able to use LLMs such as GPT-4 or Llama to generate affirmative interpretations.
Another interesting direction is to investigate the effect of affirmative interpretations on other NLU tasks, 
such as sentiment analysis or text classification.
Finally, it would be interesting to investigate the effect of affirmative interpretations on other languages, 
especially those with different word order or negation structures.

\section*{Limitations}

The scope of this paper is limited to question answering (CondaQA)
and natural language understanding (five tasks and corpora) in English with an emphasis on negation.
We leave for future work the task of exploring whether affirmative interpretations are beneficial in other languages. We acknowledge that this strategy might not generalize to other languages.

We also acknowledge that we did not conduct experiments with the latest GPT models or spend substantial amounts of time engineering prompts.
We note, however, that good faith efforts using prompts showed that ChatGPT may not be well suited for generating affirmative interpretations at this time~(Appendix \ref{sec:chatgpt}).

It is worth pointing out 
that writing affirmative interpretations for negated sentences 
might not be straightforward or even possible in some cases.
In this paper, 
we did not focus on the task of determining whether a sentence can be paraphrased without negation. 
We leave this for future work.

None of the corpora that we work with include information about the scope and focus of negation.
Therefore, we do not have any insight into the relation between affirmative interpretations and the scope and focus of a negation.

\section*{Ethics Statement}
The work in this paper does not involve human subjects.
We only use publicly available datasets and models.
We do not collect any personal information.
Therefore, this work does not raise any ethical concerns.

\section*{Acknowledgements} 
This material is based upon work supported by the National Science Foundation under Grant No. 2310334. 
Any opinions, findings, conclusions or recommendations expressed in this material 
are those of the authors and do not necessarily reflect the views of the NSF. 

We used computational resources available at the Chameleon testbed to run our experiments \citep{keahey2020lessons}. 
We are also grateful to the anonymous reviewers for their valuable comments.

\bibliography{custom}

\begin{thebibliography}{55}
\expandafter\ifx\csname natexlab\endcsname\relax\def\natexlab#1{#1}\fi

\bibitem[{Bisk et~al.(2019)Bisk, Zellers, Bras, Gao, and Choi}]{bisk2019piqa}
Yonatan Bisk, Rowan Zellers, Ronan~Le Bras, Jianfeng Gao, and Yejin Choi. 2019.
\newblock \href {http://arxiv.org/abs/1911.11641} {Piqa: Reasoning about physical commonsense in natural language}.

\bibitem[{Brown et~al.(2020)Brown, Mann, Ryder, Subbiah, Kaplan, Dhariwal, Neelakantan, Shyam, Sastry, Askell, Agarwal, Herbert-Voss, Krueger, Henighan, Child, Ramesh, Ziegler, Wu, Winter, Hesse, Chen, Sigler, Litwin, Gray, Chess, Clark, Berner, McCandlish, Radford, Sutskever, and Amodei}]{NEURIPS2020_1457c0d6}
Tom Brown, Benjamin Mann, Nick Ryder, Melanie Subbiah, Jared~D Kaplan, Prafulla Dhariwal, Arvind Neelakantan, Pranav Shyam, Girish Sastry, Amanda Askell, Sandhini Agarwal, Ariel Herbert-Voss, Gretchen Krueger, Tom Henighan, Rewon Child, Aditya Ramesh, Daniel Ziegler, Jeffrey Wu, Clemens Winter, Chris Hesse, Mark Chen, Eric Sigler, Mateusz Litwin, Scott Gray, Benjamin Chess, Jack Clark, Christopher Berner, Sam McCandlish, Alec Radford, Ilya Sutskever, and Dario Amodei. 2020.
\newblock \href {https://proceedings.neurips.cc/paper_files/paper/2020/file/1457c0d6bfcb4967418bfb8ac142f64a-Paper.pdf} {Language models are few-shot learners}.
\newblock In \emph{Advances in Neural Information Processing Systems}, volume~33, pages 1877--1901. Curran Associates, Inc.

\bibitem[{Cer et~al.(2017)Cer, Diab, Agirre, Lopez-Gazpio, and Specia}]{cer-etal-2017-semeval}
Daniel Cer, Mona Diab, Eneko Agirre, I{\~n}igo Lopez-Gazpio, and Lucia Specia. 2017.
\newblock \href {https://doi.org/10.18653/v1/S17-2001} {{S}em{E}val-2017 task 1: Semantic textual similarity multilingual and crosslingual focused evaluation}.
\newblock In \emph{Proceedings of the 11th International Workshop on Semantic Evaluation ({S}em{E}val-2017)}, pages 1--14, Vancouver, Canada. Association for Computational Linguistics.

\bibitem[{Clark et~al.(2019)Clark, Lee, Chang, Kwiatkowski, Collins, and Toutanova}]{clark-etal-2019-boolq}
Christopher Clark, Kenton Lee, Ming-Wei Chang, Tom Kwiatkowski, Michael Collins, and Kristina Toutanova. 2019.
\newblock \href {https://doi.org/10.18653/v1/N19-1300} {{B}ool{Q}: Exploring the surprising difficulty of natural yes/no questions}.
\newblock In \emph{Proceedings of the 2019 Conference of the North {A}merican Chapter of the Association for Computational Linguistics: Human Language Technologies, Volume 1 (Long and Short Papers)}, pages 2924--2936, Minneapolis, Minnesota. Association for Computational Linguistics.

\bibitem[{Clark et~al.(2018)Clark, Cowhey, Etzioni, Khot, Sabharwal, Schoenick, and Tafjord}]{clark2018think}
Peter Clark, Isaac Cowhey, Oren Etzioni, Tushar Khot, Ashish Sabharwal, Carissa Schoenick, and Oyvind Tafjord. 2018.
\newblock \href {http://arxiv.org/abs/1803.05457} {Think you have solved question answering? try arc, the ai2 reasoning challenge}.

\bibitem[{Dasigi et~al.(2019)Dasigi, Liu, Marasovi{\'c}, Smith, and Gardner}]{dasigi-etal-2019-quoref}
Pradeep Dasigi, Nelson~F. Liu, Ana Marasovi{\'c}, Noah~A. Smith, and Matt Gardner. 2019.
\newblock \href {https://doi.org/10.18653/v1/D19-1606} {{Q}uoref: A reading comprehension dataset with questions requiring coreferential reasoning}.
\newblock In \emph{Proceedings of the 2019 Conference on Empirical Methods in Natural Language Processing and the 9th International Joint Conference on Natural Language Processing (EMNLP-IJCNLP)}, pages 5925--5932, Hong Kong, China. Association for Computational Linguistics.

\bibitem[{Devlin et~al.(2019)Devlin, Chang, Lee, and Toutanova}]{devlin-etal-2019-bert}
Jacob Devlin, Ming-Wei Chang, Kenton Lee, and Kristina Toutanova. 2019.
\newblock \href {https://doi.org/10.18653/v1/N19-1423} {{BERT}: Pre-training of deep bidirectional transformers for language understanding}.
\newblock In \emph{Proceedings of the 2019 Conference of the North {A}merican Chapter of the Association for Computational Linguistics: Human Language Technologies, Volume 1 (Long and Short Papers)}, pages 4171--4186, Minneapolis, Minnesota. Association for Computational Linguistics.

\bibitem[{Dua et~al.(2019)Dua, Wang, Dasigi, Stanovsky, Singh, and Gardner}]{dua-etal-2019-drop}
Dheeru Dua, Yizhong Wang, Pradeep Dasigi, Gabriel Stanovsky, Sameer Singh, and Matt Gardner. 2019.
\newblock \href {https://doi.org/10.18653/v1/N19-1246} {{DROP}: A reading comprehension benchmark requiring discrete reasoning over paragraphs}.
\newblock In \emph{Proceedings of the 2019 Conference of the North {A}merican Chapter of the Association for Computational Linguistics: Human Language Technologies, Volume 1 (Long and Short Papers)}, pages 2368--2378, Minneapolis, Minnesota. Association for Computational Linguistics.

\bibitem[{Ettinger(2020)}]{ettinger-2020-bert}
Allyson Ettinger. 2020.
\newblock \href {https://doi.org/10.1162/tacl_a_00298} {What {BERT} is not: Lessons from a new suite of psycholinguistic diagnostics for language models}.
\newblock \emph{Transactions of the Association for Computational Linguistics}, 8:34--48.

\bibitem[{Horn(1989)}]{Horn1989-HORANH}
Laurence~R. Horn. 1989.
\newblock \emph{A Natural History of Negation}.
\newblock University of Chicago Press.

\bibitem[{Hossain and Blanco(2022)}]{hossain-blanco-2022-leveraging}
Md~Mosharaf Hossain and Eduardo Blanco. 2022.
\newblock \href {https://doi.org/10.18653/v1/2022.emnlp-main.393} {Leveraging affirmative interpretations from negation improves natural language understanding}.
\newblock In \emph{Proceedings of the 2022 Conference on Empirical Methods in Natural Language Processing}, pages 5833--5847, Abu Dhabi, United Arab Emirates. Association for Computational Linguistics.

\bibitem[{Hossain et~al.(2022{\natexlab{a}})Hossain, Chinnappa, and Blanco}]{hossain-etal-2022-analysis}
Md~Mosharaf Hossain, Dhivya Chinnappa, and Eduardo Blanco. 2022{\natexlab{a}}.
\newblock \href {https://doi.org/10.18653/v1/2022.acl-short.81} {An analysis of negation in natural language understanding corpora}.
\newblock In \emph{Proceedings of the 60th Annual Meeting of the Association for Computational Linguistics (Volume 2: Short Papers)}, pages 716--723, Dublin, Ireland. Association for Computational Linguistics.

\bibitem[{Hossain et~al.(2022{\natexlab{b}})Hossain, Holman, Kakileti, Kao, Brito, Mathews, and Blanco}]{hossain-etal-2022-question}
Md~Mosharaf Hossain, Luke Holman, Anusha Kakileti, Tiffany Kao, Nathan Brito, Aaron Mathews, and Eduardo Blanco. 2022{\natexlab{b}}.
\newblock \href {https://doi.org/10.18653/v1/2022.findings-naacl.37} {A question-answer driven approach to reveal affirmative interpretations from verbal negations}.
\newblock In \emph{Findings of the Association for Computational Linguistics: NAACL 2022}, pages 490--503, Seattle, United States. Association for Computational Linguistics.

\bibitem[{Hosseini et~al.(2021)Hosseini, Reddy, Bahdanau, Hjelm, Sordoni, and Courville}]{hosseini-etal-2021-understanding}
Arian Hosseini, Siva Reddy, Dzmitry Bahdanau, R~Devon Hjelm, Alessandro Sordoni, and Aaron Courville. 2021.
\newblock \href {https://doi.org/10.18653/v1/2021.naacl-main.102} {Understanding by understanding not: Modeling negation in language models}.
\newblock In \emph{Proceedings of the 2021 Conference of the North American Chapter of the Association for Computational Linguistics: Human Language Technologies}, pages 1301--1312, Online. Association for Computational Linguistics.

\bibitem[{Jang et~al.(2022)Jang, Mtumbuka, and Lukasiewicz}]{jang-etal-2022-beyond}
Myeongjun Jang, Frank Mtumbuka, and Thomas Lukasiewicz. 2022.
\newblock \href {https://doi.org/10.18653/v1/2022.findings-naacl.156} {Beyond distributional hypothesis: Let language models learn meaning-text correspondence}.
\newblock In \emph{Findings of the Association for Computational Linguistics: NAACL 2022}, pages 2030--2042, Seattle, United States. Association for Computational Linguistics.

\bibitem[{Kassner and Sch{\"u}tze(2020)}]{kassner-schutze-2020-negated}
Nora Kassner and Hinrich Sch{\"u}tze. 2020.
\newblock \href {https://doi.org/10.18653/v1/2020.acl-main.698} {Negated and misprimed probes for pretrained language models: Birds can talk, but cannot fly}.
\newblock In \emph{Proceedings of the 58th Annual Meeting of the Association for Computational Linguistics}, pages 7811--7818, Online. Association for Computational Linguistics.

\bibitem[{Keahey et~al.(2020)Keahey, Anderson, Zhen, Riteau, Ruth, Stanzione, Cevik, Colleran, Gunawi, Hammock, Mambretti, Barnes, Halbach, Rocha, and Stubbs}]{keahey2020lessons}
Kate Keahey, Jason Anderson, Zhuo Zhen, Pierre Riteau, Paul Ruth, Dan Stanzione, Mert Cevik, Jacob Colleran, Haryadi~S. Gunawi, Cody Hammock, Joe Mambretti, Alexander Barnes, Fran\c{c}ois Halbach, Alex Rocha, and Joe Stubbs. 2020.
\newblock Lessons learned from the chameleon testbed.
\newblock In \emph{Proceedings of the 2020 USENIX Annual Technical Conference (USENIX ATC '20)}. USENIX Association.

\bibitem[{Khandelwal and Sawant(2020)}]{khandelwal-sawant-2020-negbert}
Aditya Khandelwal and Suraj Sawant. 2020.
\newblock \href {https://aclanthology.org/2020.lrec-1.704} {{N}eg{BERT}: A transfer learning approach for negation detection and scope resolution}.
\newblock In \emph{Proceedings of the Twelfth Language Resources and Evaluation Conference}, pages 5739--5748, Marseille, France. European Language Resources Association.

\bibitem[{Khashabi et~al.(2018)Khashabi, Chaturvedi, Roth, Upadhyay, and Roth}]{khashabi-etal-2018-looking}
Daniel Khashabi, Snigdha Chaturvedi, Michael Roth, Shyam Upadhyay, and Dan Roth. 2018.
\newblock \href {https://doi.org/10.18653/v1/N18-1023} {Looking beyond the surface: A challenge set for reading comprehension over multiple sentences}.
\newblock In \emph{Proceedings of the 2018 Conference of the North {A}merican Chapter of the Association for Computational Linguistics: Human Language Technologies, Volume 1 (Long Papers)}, pages 252--262, New Orleans, Louisiana. Association for Computational Linguistics.

\bibitem[{Khashabi et~al.(2020)Khashabi, Khot, and Sabharwal}]{khashabi-etal-2020-bang}
Daniel Khashabi, Tushar Khot, and Ashish Sabharwal. 2020.
\newblock \href {https://doi.org/10.18653/v1/2020.emnlp-main.12} {More bang for your buck: Natural perturbation for robust question answering}.
\newblock In \emph{Proceedings of the 2020 Conference on Empirical Methods in Natural Language Processing (EMNLP)}, pages 163--170, Online. Association for Computational Linguistics.

\bibitem[{Khashabi et~al.(2022)Khashabi, Kordi, and Hajishirzi}]{khashabi2022unifiedqav2}
Daniel Khashabi, Yeganeh Kordi, and Hannaneh Hajishirzi. 2022.
\newblock \href {http://arxiv.org/abs/2202.12359} {Unifiedqa-v2: Stronger generalization via broader cross-format training}.

\bibitem[{Khot et~al.(2020)Khot, Clark, Guerquin, Jansen, and Sabharwal}]{Khot_Clark_Guerquin_Jansen_Sabharwal_2020}
Tushar Khot, Peter Clark, Michal Guerquin, Peter Jansen, and Ashish Sabharwal. 2020.
\newblock \href {https://doi.org/10.1609/aaai.v34i05.6319} {Qasc: A dataset for question answering via sentence composition}.
\newblock \emph{Proceedings of the AAAI Conference on Artificial Intelligence}, 34(05):8082--8090.

\bibitem[{Ko{\v{c}}isk{\'y} et~al.(2018)Ko{\v{c}}isk{\'y}, Schwarz, Blunsom, Dyer, Hermann, Melis, and Grefenstette}]{kocisky-etal-2018-narrativeqa}
Tom{\'a}{\v{s}} Ko{\v{c}}isk{\'y}, Jonathan Schwarz, Phil Blunsom, Chris Dyer, Karl~Moritz Hermann, G{\'a}bor Melis, and Edward Grefenstette. 2018.
\newblock \href {https://doi.org/10.1162/tacl_a_00023} {The {N}arrative{QA} reading comprehension challenge}.
\newblock \emph{Transactions of the Association for Computational Linguistics}, 6:317--328.

\bibitem[{Kwiatkowski et~al.(2019)Kwiatkowski, Palomaki, Redfield, Collins, Parikh, Alberti, Epstein, Polosukhin, Devlin, Lee, Toutanova, Jones, Kelcey, Chang, Dai, Uszkoreit, Le, and Petrov}]{kwiatkowski-etal-2019-natural}
Tom Kwiatkowski, Jennimaria Palomaki, Olivia Redfield, Michael Collins, Ankur Parikh, Chris Alberti, Danielle Epstein, Illia Polosukhin, Jacob Devlin, Kenton Lee, Kristina Toutanova, Llion Jones, Matthew Kelcey, Ming-Wei Chang, Andrew~M. Dai, Jakob Uszkoreit, Quoc Le, and Slav Petrov. 2019.
\newblock \href {https://doi.org/10.1162/tacl_a_00276} {Natural questions: A benchmark for question answering research}.
\newblock \emph{Transactions of the Association for Computational Linguistics}, 7:452--466.

\bibitem[{Lai et~al.(2017)Lai, Xie, Liu, Yang, and Hovy}]{lai-etal-2017-race}
Guokun Lai, Qizhe Xie, Hanxiao Liu, Yiming Yang, and Eduard Hovy. 2017.
\newblock \href {https://doi.org/10.18653/v1/D17-1082} {{RACE}: Large-scale {R}e{A}ding comprehension dataset from examinations}.
\newblock In \emph{Proceedings of the 2017 Conference on Empirical Methods in Natural Language Processing}, pages 785--794, Copenhagen, Denmark. Association for Computational Linguistics.

\bibitem[{Levesque et~al.(2012)Levesque, Davis, and Morgenstern}]{levesque_winograd_2012}
Hector~J. Levesque, Ernest Davis, and Leora Morgenstern. 2012.
\newblock \href {https://cs.nyu.edu/faculty/davise/papers/WSKR2012.pdf} {The {Winograd} {Schema} {Challenge}}.
\newblock In \emph{Proceedings of the {Thirteenth} {International} {Conference} on {Principles} of {Knowledge} {Representation} and {Reasoning}}, {KR}'12, pages 552--561. AAAI Press, Rome, Italy.

\bibitem[{Lin et~al.(2019)Lin, Tafjord, Clark, and Gardner}]{lin-etal-2019-reasoning}
Kevin Lin, Oyvind Tafjord, Peter Clark, and Matt Gardner. 2019.
\newblock \href {https://doi.org/10.18653/v1/D19-5808} {Reasoning over paragraph effects in situations}.
\newblock In \emph{Proceedings of the 2nd Workshop on Machine Reading for Question Answering}, pages 58--62, Hong Kong, China. Association for Computational Linguistics.

\bibitem[{Liu et~al.(2019)Liu, Ott, Goyal, Du, Joshi, Chen, Levy, Lewis, Zettlemoyer, and Stoyanov}]{liu2019roberta}
Yinhan Liu, Myle Ott, Naman Goyal, Jingfei Du, Mandar Joshi, Danqi Chen, Omer Levy, Mike Lewis, Luke Zettlemoyer, and Veselin Stoyanov. 2019.
\newblock \href {http://arxiv.org/abs/1907.11692} {Roberta: A robustly optimized bert pretraining approach}.

\bibitem[{McNemar(1947)}]{mcnemar1947note}
Quinn McNemar. 1947.
\newblock Note on the sampling error of the difference between correlated proportions or percentages.
\newblock \emph{Psychometrika}, 12(2):153--157.

\bibitem[{Mihaylov et~al.(2018)Mihaylov, Clark, Khot, and Sabharwal}]{mihaylov-etal-2018-suit}
Todor Mihaylov, Peter Clark, Tushar Khot, and Ashish Sabharwal. 2018.
\newblock \href {https://doi.org/10.18653/v1/D18-1260} {Can a suit of armor conduct electricity? a new dataset for open book question answering}.
\newblock In \emph{Proceedings of the 2018 Conference on Empirical Methods in Natural Language Processing}, pages 2381--2391, Brussels, Belgium. Association for Computational Linguistics.

\bibitem[{Morante and Daelemans(2012)}]{morante-daelemans-2012-conandoyle}
Roser Morante and Walter Daelemans. 2012.
\newblock \href {http://www.lrec-conf.org/proceedings/lrec2012/pdf/221_Paper.pdf} {{C}onan{D}oyle-neg: Annotation of negation cues and their scope in conan doyle stories}.
\newblock In \emph{Proceedings of the Eighth International Conference on Language Resources and Evaluation ({LREC}'12)}, pages 1563--1568, Istanbul, Turkey. European Language Resources Association (ELRA).

\bibitem[{Morante et~al.(2011)Morante, Schrauwen, and Daelemans}]{Morante2011NegationCues}
Roser Morante, Sarah Schrauwen, and Walter Daelemans. 2011.
\newblock Annotation of negation cues and their scope: Guidelines v1.
\newblock Technical Report CTRS-003, Computational Linguistics and Psycholinguistics Technical Report Series.

\bibitem[{Ouyang et~al.(2022)Ouyang, Wu, Jiang, Almeida, Wainwright, Mishkin, Zhang, Agarwal, Slama, Ray, Schulman, Hilton, Kelton, Miller, Simens, Askell, Welinder, Christiano, Leike, and Lowe}]{ouyang2022training}
Long Ouyang, Jeff Wu, Xu~Jiang, Diogo Almeida, Carroll~L. Wainwright, Pamela Mishkin, Chong Zhang, Sandhini Agarwal, Katarina Slama, Alex Ray, John Schulman, Jacob Hilton, Fraser Kelton, Luke Miller, Maddie Simens, Amanda Askell, Peter Welinder, Paul Christiano, Jan Leike, and Ryan Lowe. 2022.
\newblock \href {http://arxiv.org/abs/2203.02155} {Training language models to follow instructions with human feedback}.

\bibitem[{Phang et~al.(2020)Phang, Yeres, Swanson, Liu, Tenney, Htut, Vania, Wang, and Bowman}]{phang2020jiant}
Jason Phang, Phil Yeres, Jesse Swanson, Haokun Liu, Ian~F. Tenney, Phu~Mon Htut, Clara Vania, Alex Wang, and Samuel~R. Bowman. 2020.
\newblock \texttt{jiant} 2.0: A software toolkit for research on general-purpose text understanding models.
\newblock \url{http://jiant.info/}.

\bibitem[{Pilehvar and Camacho-Collados(2019)}]{pilehvar-camacho-collados-2019-wic}
Mohammad~Taher Pilehvar and Jose Camacho-Collados. 2019.
\newblock \href {https://doi.org/10.18653/v1/N19-1128} {{W}i{C}: the word-in-context dataset for evaluating context-sensitive meaning representations}.
\newblock In \emph{Proceedings of the 2019 Conference of the North {A}merican Chapter of the Association for Computational Linguistics: Human Language Technologies, Volume 1 (Long and Short Papers)}, pages 1267--1273, Minneapolis, Minnesota. Association for Computational Linguistics.

\bibitem[{Raffel et~al.(2020)Raffel, Shazeer, Roberts, Lee, Narang, Matena, Zhou, Li, and Liu}]{t5paper}
Colin Raffel, Noam Shazeer, Adam Roberts, Katherine Lee, Sharan Narang, Michael Matena, Yanqi Zhou, Wei Li, and Peter~J. Liu. 2020.
\newblock \href {http://jmlr.org/papers/v21/20-074.html} {Exploring the limits of transfer learning with a unified text-to-text transformer}.
\newblock \emph{Journal of Machine Learning Research}, 21(140):1--67.

\bibitem[{Rajpurkar et~al.(2018)Rajpurkar, Jia, and Liang}]{rajpurkar-etal-2018-know}
Pranav Rajpurkar, Robin Jia, and Percy Liang. 2018.
\newblock \href {https://doi.org/10.18653/v1/P18-2124} {Know what you don{'}t know: Unanswerable questions for {SQ}u{AD}}.
\newblock In \emph{Proceedings of the 56th Annual Meeting of the Association for Computational Linguistics (Volume 2: Short Papers)}, pages 784--789, Melbourne, Australia. Association for Computational Linguistics.

\bibitem[{Rajpurkar et~al.(2016)Rajpurkar, Zhang, Lopyrev, and Liang}]{rajpurkar-etal-2016-squad}
Pranav Rajpurkar, Jian Zhang, Konstantin Lopyrev, and Percy Liang. 2016.
\newblock \href {https://doi.org/10.18653/v1/D16-1264} {{SQ}u{AD}: 100,000+ questions for machine comprehension of text}.
\newblock In \emph{Proceedings of the 2016 Conference on Empirical Methods in Natural Language Processing}, pages 2383--2392, Austin, Texas. Association for Computational Linguistics.

\bibitem[{Ravichander et~al.(2022)Ravichander, Gardner, and Marasovic}]{ravichander-etal-2022-condaqa}
Abhilasha Ravichander, Matt Gardner, and Ana Marasovic. 2022.
\newblock \href {https://doi.org/10.18653/v1/2022.emnlp-main.598} {{CONDAQA}: A contrastive reading comprehension dataset for reasoning about negation}.
\newblock In \emph{Proceedings of the 2022 Conference on Empirical Methods in Natural Language Processing}, pages 8729--8755, Abu Dhabi, United Arab Emirates. Association for Computational Linguistics.

\bibitem[{Richardson et~al.(2013)Richardson, Burges, and Renshaw}]{richardson-etal-2013-mctest}
Matthew Richardson, Christopher~J.C. Burges, and Erin Renshaw. 2013.
\newblock \href {https://aclanthology.org/D13-1020} {{MCT}est: A challenge dataset for the open-domain machine comprehension of text}.
\newblock In \emph{Proceedings of the 2013 Conference on Empirical Methods in Natural Language Processing}, pages 193--203, Seattle, Washington, USA. Association for Computational Linguistics.

\bibitem[{Roemmele et~al.(2011)Roemmele, Bejan, and Gordon}]{roemmele-etal-2011-choice}
Melissa Roemmele, Cosmin Bejan, and Andrew Gordon. 2011.
\newblock Choice of plausible alternatives: An evaluation of commonsense causal reasoning.

\bibitem[{Sakaguchi et~al.(2020)Sakaguchi, Le~Bras, Bhagavatula, and Choi}]{Sakaguchi_LeBras_Bhagavatula_Choi_2020}
Keisuke Sakaguchi, Ronan Le~Bras, Chandra Bhagavatula, and Yejin Choi. 2020.
\newblock \href {https://doi.org/10.1609/aaai.v34i05.6399} {Winogrande: An adversarial winograd schema challenge at scale}.
\newblock \emph{Proceedings of the AAAI Conference on Artificial Intelligence}, 34(05):8732--8740.

\bibitem[{Sap et~al.(2019)Sap, Rashkin, Chen, Le~Bras, and Choi}]{sap-etal-2019-social}
Maarten Sap, Hannah Rashkin, Derek Chen, Ronan Le~Bras, and Yejin Choi. 2019.
\newblock \href {https://doi.org/10.18653/v1/D19-1454} {Social {IQ}a: Commonsense reasoning about social interactions}.
\newblock In \emph{Proceedings of the 2019 Conference on Empirical Methods in Natural Language Processing and the 9th International Joint Conference on Natural Language Processing (EMNLP-IJCNLP)}, pages 4463--4473, Hong Kong, China. Association for Computational Linguistics.

\bibitem[{Sarabi et~al.(2019)Sarabi, Killian, Blanco, and Palmer}]{sarabi-etal-2019-corpus}
Zahra Sarabi, Erin Killian, Eduardo Blanco, and Alexis Palmer. 2019.
\newblock \href {https://doi.org/10.18653/v1/S19-1017} {A corpus of negations and their underlying positive interpretations}.
\newblock In \emph{Proceedings of the Eighth Joint Conference on Lexical and Computational Semantics (*{SEM} 2019)}, pages 158--167, Minneapolis, Minnesota. Association for Computational Linguistics.

\bibitem[{Singh et~al.(2023)Singh, Kumar, and Sridhar}]{singh-etal-2023-nlms}
Rituraj Singh, Rahul Kumar, and Vivek Sridhar. 2023.
\newblock \href {https://doi.org/10.18653/v1/2023.findings-emnlp.873} {{NLM}s: Augmenting negation in language models}.
\newblock In \emph{Findings of the Association for Computational Linguistics: EMNLP 2023}, pages 13104--13116, Singapore. Association for Computational Linguistics.

\bibitem[{Socher et~al.(2013)Socher, Perelygin, Wu, Chuang, Manning, Ng, and Potts}]{socher-etal-2013-recursive}
Richard Socher, Alex Perelygin, Jean Wu, Jason Chuang, Christopher~D. Manning, Andrew Ng, and Christopher Potts. 2013.
\newblock \href {https://aclanthology.org/D13-1170} {Recursive deep models for semantic compositionality over a sentiment treebank}.
\newblock In \emph{Proceedings of the 2013 Conference on Empirical Methods in Natural Language Processing}, pages 1631--1642, Seattle, Washington, USA. Association for Computational Linguistics.

\bibitem[{Talmor et~al.(2019)Talmor, Herzig, Lourie, and Berant}]{talmor-etal-2019-commonsenseqa}
Alon Talmor, Jonathan Herzig, Nicholas Lourie, and Jonathan Berant. 2019.
\newblock \href {https://doi.org/10.18653/v1/N19-1421} {{C}ommonsense{QA}: A question answering challenge targeting commonsense knowledge}.
\newblock In \emph{Proceedings of the 2019 Conference of the North {A}merican Chapter of the Association for Computational Linguistics: Human Language Technologies, Volume 1 (Long and Short Papers)}, pages 4149--4158, Minneapolis, Minnesota. Association for Computational Linguistics.

\bibitem[{Trischler et~al.(2017)Trischler, Wang, Yuan, Harris, Sordoni, Bachman, and Suleman}]{trischler-etal-2017-newsqa}
Adam Trischler, Tong Wang, Xingdi Yuan, Justin Harris, Alessandro Sordoni, Philip Bachman, and Kaheer Suleman. 2017.
\newblock \href {https://doi.org/10.18653/v1/W17-2623} {{N}ews{QA}: A machine comprehension dataset}.
\newblock In \emph{Proceedings of the 2nd Workshop on Representation Learning for {NLP}}, pages 191--200, Vancouver, Canada. Association for Computational Linguistics.

\bibitem[{Truong et~al.(2022)Truong, Baldwin, Cohn, and Verspoor}]{truong-etal-2022-improving}
Thinh Truong, Timothy Baldwin, Trevor Cohn, and Karin Verspoor. 2022.
\newblock \href {https://doi.org/10.18653/v1/2022.naacl-main.309} {Improving negation detection with negation-focused pre-training}.
\newblock In \emph{Proceedings of the 2022 Conference of the North American Chapter of the Association for Computational Linguistics: Human Language Technologies}, pages 4188--4193, Seattle, United States. Association for Computational Linguistics.

\bibitem[{Truong et~al.(2023)Truong, Baldwin, Verspoor, and Cohn}]{truong-etal-2023-language}
Thinh~Hung Truong, Timothy Baldwin, Karin Verspoor, and Trevor Cohn. 2023.
\newblock \href {https://doi.org/10.18653/v1/2023.starsem-1.10} {Language models are not naysayers: an analysis of language models on negation benchmarks}.
\newblock In \emph{Proceedings of the 12th Joint Conference on Lexical and Computational Semantics (*SEM 2023)}, pages 101--114, Toronto, Canada. Association for Computational Linguistics.

\bibitem[{van Son et~al.(2016)van Son, van Miltenburg, and Morante}]{van-son-etal-2016-building}
Chantal van Son, Emiel van Miltenburg, and Roser Morante. 2016.
\newblock \href {https://aclanthology.org/W16-5007} {Building a dictionary of affixal negations}.
\newblock In \emph{Proceedings of the Workshop on Extra-Propositional Aspects of Meaning in Computational Linguistics ({E}x{P}ro{M})}, pages 49--56, Osaka, Japan. The COLING 2016 Organizing Committee.

\bibitem[{Vorobev and Kuznetsov(2023)}]{chatgpt_paraphraser}
Vladimir Vorobev and Maxim Kuznetsov. 2023.
\newblock \href {https://huggingface.co/humarin/chatgpt_paraphraser_on_T5_base} {A paraphrasing model based on chatgpt paraphrases}.

\bibitem[{Wang et~al.(2019)Wang, Pruksachatkun, Nangia, Singh, Michael, Hill, Levy, and Bowman}]{superglue}
Alex Wang, Yada Pruksachatkun, Nikita Nangia, Amanpreet Singh, Julian Michael, Felix Hill, Omer Levy, and Samuel~R. Bowman. 2019.
\newblock \href {http://arxiv.org/abs/1905.00537} {Superglue: {A} stickier benchmark for general-purpose language understanding systems}.
\newblock \emph{CoRR}, abs/1905.00537.

\bibitem[{Wang et~al.(2018)Wang, Singh, Michael, Hill, Levy, and Bowman}]{wang-etal-2018-glue}
Alex Wang, Amanpreet Singh, Julian Michael, Felix Hill, Omer Levy, and Samuel Bowman. 2018.
\newblock \href {https://doi.org/10.18653/v1/W18-5446} {{GLUE}: A multi-task benchmark and analysis platform for natural language understanding}.
\newblock In \emph{Proceedings of the 2018 {EMNLP} Workshop {B}lackbox{NLP}: Analyzing and Interpreting Neural Networks for {NLP}}, pages 353--355, Brussels, Belgium. Association for Computational Linguistics.

\bibitem[{Wolf et~al.(2020)Wolf, Debut, Sanh, Chaumond, Delangue, Moi, Cistac, Rault, Louf, Funtowicz, Davison, Shleifer, von Platen, Ma, Jernite, Plu, Xu, Le~Scao, Gugger, Drame, Lhoest, and Rush}]{wolf-etal-2020-transformers}
Thomas Wolf, Lysandre Debut, Victor Sanh, Julien Chaumond, Clement Delangue, Anthony Moi, Pierric Cistac, Tim Rault, Remi Louf, Morgan Funtowicz, Joe Davison, Sam Shleifer, Patrick von Platen, Clara Ma, Yacine Jernite, Julien Plu, Canwen Xu, Teven Le~Scao, Sylvain Gugger, Mariama Drame, Quentin Lhoest, and Alexander Rush. 2020.
\newblock \href {https://doi.org/10.18653/v1/2020.emnlp-demos.6} {Transformers: State-of-the-art natural language processing}.
\newblock In \emph{Proceedings of the 2020 Conference on Empirical Methods in Natural Language Processing: System Demonstrations}, pages 38--45, Online. Association for Computational Linguistics.

\end{thebibliography}

\appendix

\section{Paraphrases vs. Affirmative Interpretations} 
\label{sec:affirmativeexamples}

\begin{table*}
\small
\centering
  \begin{tabular}{l c c}
    \toprule
    & Negation? & Same Meaning? \\
    Original Sentence with Negation: \\
    ~~The lightning strikes caused no serious permanent damage. & Yes & n/a \\
    \midrule
    Automatically Generated Paraphrases (unfiltered): \\ 
    ~~The lightning did not cause any damage. & Yes & No \\
    ~~The lightning did not cause any significant and permanent damage. & Yes & Yes \\
    ~~The lightning strikes caused serious permanent damage. & No & No \\
    ~~Lightning strikes caused short-term damage. & No & Yes \\
    \bottomrule
    \end{tabular}
    \caption{
      \label{fig:affirmativeexamples}
      Examples of automatically generated paraphrases from a negated sentence.
      The first two paraphrases contain negation, and only the second one preserves meaning.
      The next two paraphrases do not contain negation, and only the fourth one preserves meaning.
      Only the fourth automatically obtained paraphrase is an affirmative interpretation:
      it does not contain negation and it is a true paraphrase of the original sentence with negation---\emph{not causing serious permanent damage} carries roughly the same meaning than \emph{causing short-term damage}.
    }
\end{table*}

Affirmative interpretations are paraphrases without negation.
Table~\ref{fig:affirmativeexamples} shows examples of automatically generated paraphrases from a negated sentence.
Not all of them are correct affirmative interpretations: some
(a)~contain negation or
(b)~do not preserve the meaning of the original sentence with negation (and thus they are not actual paraphrases to begin with).
The definition of affirmative interpretation is a paraphrase (i.e., rewording that preserves meaning) not containing negation.

Note that an automatically obtained paraphrase that does not preserve the full meaning (and thus does not satisfy the definition of affirmative interpretation) does not necessarily contradict the meaning of the original sentence with negation.
For example, \emph{I stayed home today} is not a true paraphrase of \emph{I didn't go shopping today} but is not a contradiction either.
In this example, obtaining \emph{I stayed home today}, despite being only plausible and not a paraphrase of \emph{I didn't go shopping today},
could be useful to answer questions such as ``Did I go shopping today?'' as \emph{staying home} contradicts \emph{going shopping}. 

\section{NLU Corpora}
\label{sec:nlucorpora}

\begin{table*}
\small
\centering
    \begin{tabularx}{\textwidth}{lXp{1.25in}} 
      \toprule
      & Input & Output \\
      \midrule
      Question Answering \\
      ~~~~CommonsenseQA & What are you waiting alongside with when you're in a reception area? & D \\
      & A) Motel, B) Chair, C) Hospital, D) People, E) Hotel \\
      \midrule
      Similarity and Paraphrasing \\
      ~~~~STS-B & Three men are playing guitars. & 3.75 (out of 5)\\
      & Three men are on stage playing guitars. \\
      \midrule
      Inference \\
      ~~~~QNLI & What happened to Dane? & \multirow{2}{1.25in}{Entailment (i.e., question is answered)} \\
      & Dane was killed in a horse-riding accident when Nikola was five. \\
      \midrule
      Word Sense Disambiguation \\
      ~~~~WiC & Room and \emph{board}. & Not same meaning \\
      & He nailed \emph{boards} across the windows. \\
      \midrule
      Coreference Resolution \\
      ~~~~WSC & Mark told \emph{Pete} many lies about himself, which Pete included in his book. \emph{He} should have been
      more truthful. & Not coreferent \\
      \bottomrule
      \end{tabularx}
      \caption{
        \label{tab:nlucorpora}
        Examples of instances from the NLU tasks used in our experiments.
        The first column indicates the task and the corpus.
        The second column shows the input to the system. 
        The third column shows the expected output.
      }
\end{table*}

Table~\ref{tab:nlucorpora} shows examples from the five NLU corpora that we experiment with.
The examples are from the development set of each corpus.
In our experiments,
we append the affirmative interpretation of the negated sentence in the input 
to the end of the input after a special token.

\section{Attempting to Generate Affirmative Interpretations with ChatGPT}
\label{sec:chatgpt}

\begin{figure}
  \includegraphics[width=\columnwidth]{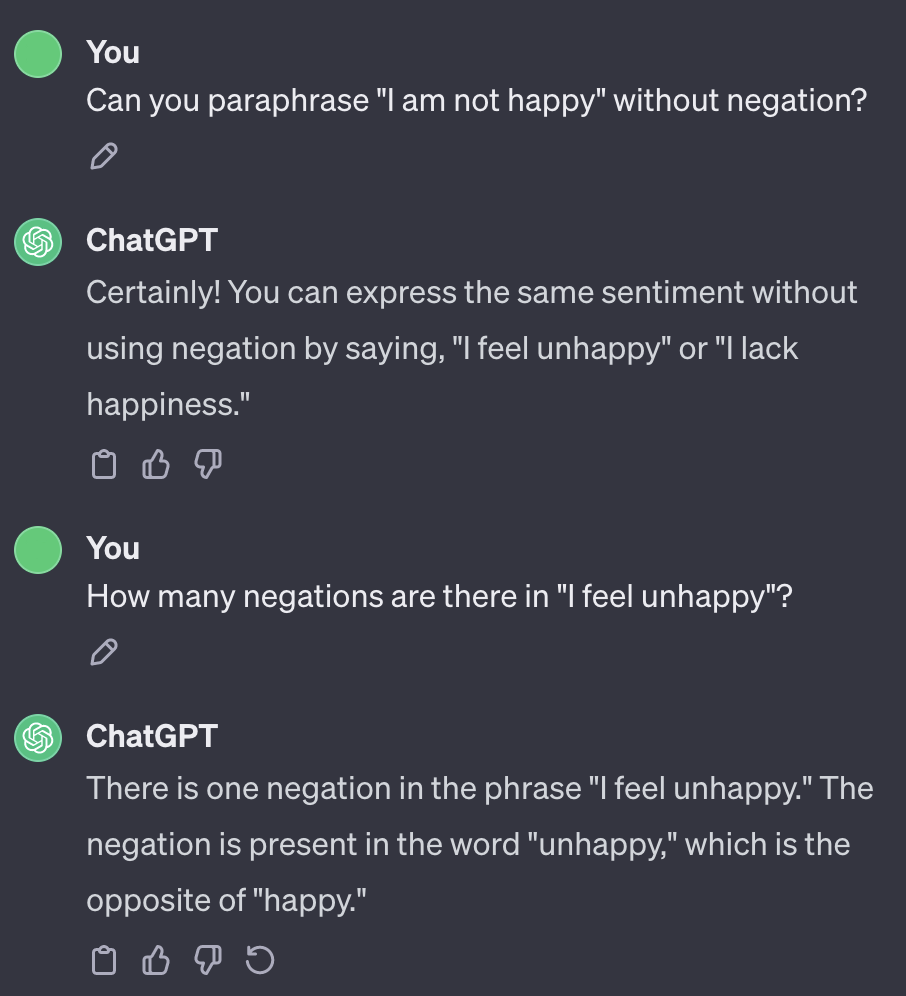}
  \caption{Attempting to generate affirmative interpretations with ChatGPT results in a nonsensical conversation.
    ChatGPT appears to be able to identify negations yet uses them when instructed to not do so}
  \label{f:affirmative_chatgpt}
  \end{figure}

At the time of writing, ChatGPT cannot reliably generate affirmative interpretations (i.e., paraphrase without using negation).
In the example in Figure \ref{f:affirmative_chatgpt},
it appears convinced to be able to do so, yet it clearly fails:
\emph{unhappy} and \emph{lack} are negations.
Perhaps surprisingly, ChatGPT appears to know that the generated output does contain negation.\\

\section{Affirmative Interpretations Examples} 
\label{sec:afanalysis}

\begin{table*}
\small
  \begin{tabularx}{\textwidth}{lXc}
    \toprule
      \multicolumn{2}{l}{Negated Sentence and Affirmative Interpretations} & Correct? \\ \midrule
      {Negated Sentence} & The National Palace is one of Managua's oldest buildings, undamaged by the 1972 earthquake. & n/a  \\ \addlinespace
      {\texttt{A\textsubscript{HB}}} & The National Palace, one of Managua's oldest buildings, survived the 1972 earthquake. & Yes \\
      {\texttt{A\textsubscript{CG}}} & The National Palace, which was one of the oldest structures in Managua, remained intact following the 1972 earthquake. & Yes \\
    \midrule
      {Negated Sentence} & It is not rare to find pearls that measure as much as 14mm across. & n/a \\ \addlinespace
      {\texttt{A\textsubscript{HB}}} & It is not uncommon to find pearls that measure as much as 14mm across. & No \\
      {\texttt{A\textsubscript{CG}}} & The size of 14mm pearls is not uncommon. & No \\
      Human & It is common to find pearls that measure as much as 14mm across. & Yes \\
    \bottomrule
  \end{tabularx}
  \caption{
    \label{tab:examples}
    Examples of negated sentences and affirmative interpretations generated by \texttt{T5-HB} (\texttt{A\textsubscript{HB}}) and \texttt{T5-CG} (\texttt{A\textsubscript{CG}}).
    The last column indicates whether affirmative interpretation are correct (i.e., meaning preserving and without negation).
    \emph{Human} is a human-generated affirmative interpretation.
  }
\end{table*}

\begin{table*}
\small
  \begin{tabularx}{\textwidth}{X}
    \toprule
      - \textit{Negated Sentence}: An increasing minority of young people cannot understand Japanese and instead use the Ryukyuan languages only. \\
      - \textit{(wrong) Affirmative Interpretation}: An increasing minority of young people understand only Ryukyuan languages instead. \\
      - \textit{Observation}: The affirmative interpretations drops an important part of the meaning of the negated sentence:
        not understanding Japanese. We note, however, that the affirmative interpretation is factual given the negated sentence and it is likely to be useful for downstream tasks. \\
    \midrule
      - \textit{Negated Sentence}: By war's end, no other nation formally recognized the Confederacy. \\
      - \textit{(wrong) Affirmative Interpretation}: Only one nation recognized the Confederacy at the end of war. \\
      - \textit{Observation}: This error seems to be due to lack of context of the negated sentence.
        The affirmative interpretation does not have negation and is plausible, but it is impossible to know how many nations recognized the Confederacy 
        without additional context.
        Indeed, \emph{no other nation did X} could mean that \emph{any number of nations did X}.
     \\ \bottomrule
  \end{tabularx}
  \caption{
    \label{tab:failedexamples} 
      A couple examples of negated sentence from CondaQA and automatically generated affirmative interpretations that are wrong.
      We also provide our observations.
  }
\end{table*}

Table~\ref{tab:examples} shows two negated sentences and their automatically obtained affirmative interpretations.
The bottom half of the table includes errors, as the automatically generated affirmative interpretations contain negations.
Table~\ref{tab:failedexamples} contains a couple examples from CondaQA in which the process to generate affirmative interpretations made mistakes along with our observations.

\section{Training Details with CondaQA}
\label{sec:condaqatraining}

\begin{table}
\small
  \centering
  \newcommand{\sig}{$^{\ast}$}
\small
\begin{tabular}{l l r}
    \toprule
	\multicolumn{2}{c}{Instance Representation}             \\
	\cmidrule(lr){1-2} \multicolumn{1}{c}{Training}                     & \multicolumn{1}{c}{Testing}      & \multicolumn{1}{c}{Learning Rate}  \\
	\midrule
	P+Q                                                             & P+Q                              &   1e-5                                 \\
	P+Q+S & P+Q+S & 5e-6 \\
	P+Q+\texttt{P\textsubscript{CG}} & P+Q+\texttt{P\textsubscript{CG}} & 1e-5 \\

	P+Q+\texttt{A\textsubscript{HB}}                                & P+Q                              &  1e-5                                  \\
	P+Q+\texttt{A\textsubscript{HB}}                                & P+Q+\texttt{A\textsubscript{HB}} & 1e-4                                    \\
	P+Q+\texttt{A\textsubscript{CG}}                                & P+Q                              &    1e-5                                \\
	P+Q+\texttt{A\textsubscript{CG}}                                & P+Q+\texttt{A\textsubscript{CG}} &  1e-4                                   \\
	P+Q+\texttt{A\textsubscript{HB}}+\texttt{A\textsubscript{CG}}  & P+Q+\texttt{A\textsubscript{HB}}+\texttt{A\textsubscript{CG}} & 1e-5 \\
	P+Q+\texttt{A\textsubscript{G}}                                 & P+Q                              &  1e-5                                  \\
	P+Q+\texttt{A\textsubscript{G}}                                 & P+Q+\texttt{A\textsubscript{HB}} &  1e-5                                  \\
	P+Q+\texttt{A\textsubscript{G}}                                 & P+Q+\texttt{A\textsubscript{CG}} &    1e-5                                \\
	P+Q+\texttt{A\textsubscript{G}} or \texttt{A\textsubscript{HB}} & P+Q                              &   1e-5                                \\
	P+Q+\texttt{A\textsubscript{G}} or \texttt{A\textsubscript{HB}} & P+Q+\texttt{A\textsubscript{HB}} & 5e-5                               \\
	P+Q+\texttt{A\textsubscript{G}} or \texttt{A\textsubscript{CG}} & P+Q                              & 1e-5                                    \\
	P+Q+\texttt{A\textsubscript{G}} or \texttt{A\textsubscript{CG}} & P+Q+\texttt{A\textsubscript{CG}} & 5e-5                                    \\
	\bottomrule
\end{tabular}
  \caption{
    \label{table:condaqahyperparameters} 
    Learning rates used in our experiments with CondaQA.
    Note that \texttt{A\textsubscript{G}} affirmative interpretations are not available at testing time.
  }
\end{table}

We use the RoBERTa-Large model \cite{liu2019roberta} for our experiments with CondaQA.
We use the implementation of RoBERTa-Large in the HuggingFace Transformers library \cite{wolf-etal-2020-transformers}.
The model is trained using early stopping with a patience of 3 epochs and batch size 16.
Table~\ref{table:condaqahyperparameters} shows the learning rates that we used for our experiments with CondaQA.
We use the default values for the other hyperparameters.

\section{Training Details with Additional NLU Tasks}
\label{sec:nlutraining}
We use the implementation by
\citet{phang2020jiant}
with RoBERTa-Large as the base model.
We use the default values for the hyperparameters,
with the exception of the learning rate, batch size and maximum number of epochs for early stopping.

\begin{table*}
\small
  \centering
  \small
\begin{tabular}{l c@{\hspace{.10in}} r r r r r }
	\toprule 
 &  & \multicolumn{1}{c}{{CmmnsnsQA}} & \multicolumn{1}{c}{{STS-B}} & \multicolumn{1}{c}{{QNLI}} & \multicolumn{1}{c}{{WiC}} & \multicolumn{1}{c}{{WSC}} \\
	\\
	\midrule RoBERTa  && 1e-5 (16) & 1e-5 (16) & 1e-5 (8) & 1e-5 (16) & 1e-6 (16)\\
	\midrule \multicolumn{1}{l}{RoBERTa w/ Affirmative Interpret.}                                       \\
	~~~~obtained using \texttt{T5-HB} (\texttt{A\textsubscript{HB}}) && 5e-6 (16) & 5e-6 (8)  & 5e-6 (16)  & 1e-5 (16) & 5e-6 (16) \\
	~~~~obtained using \texttt{T5-CG} (\texttt{A\textsubscript{CG}}) && 5e-6 (16) & 5e-6 (16) & 1e-5 (16)  & 5e-6 (16) & 5e-6 (16) \\
	\bottomrule
\end{tabular}
  \caption{
    \label{table:nluhyperparameters} 
    The learning rates (and batch sizes) used in our experiments with each corpus.
  }
\end{table*}

Table~\ref{table:nluhyperparameters} shows the learning rates and batch sizes
that we used for our experiments on each corpus.

\section{CondaQA Dataset}
\label{sec:condaqaexample}

\begin{figure*}
\small
  \begin{tabularx}{\textwidth}{p{.9in}X}
    \toprule
    {Original Passage:} & A semiconductor diode is a device typically made from a single p-n junction. At the junction of a p-type and an n-type semiconductor, there forms a depletion region where current conduction is inhibited by the lack of mobile charge carriers. When the device is "forward biased" (connected with the p-side at higher electric potential than the n-side), this depletion region is diminished, allowing for significant conduction, while only very small current can be achieved when the diode is "reverse biased" and thus the depletion region expanded. \\ \addlinespace
    {Original Sentence (with Negation):} & At the junction of a p-type and an n-type semiconductor, there forms a depletion region where current conduction is inhibited by the lack of mobile charge carriers. \\ \addlinespace
    {Negation Cue:} & lack \\ \addlinespace
    {Edited Passage:} &  A semiconductor diode is a device typically made from a single p-n junction. At the junction of a p-type and an n-type semiconductor there forms a depletion region where current conduction is inhibited by the absence of mobile charge carriers. When the device is "forward biased" (connected with the p-side at higher electric potential than the n-side), this depletion region is diminished, allowing for significant conduction, while only very small current can be achieved when the diode is "reverse biased" and thus the depletion region expanded. \\ \addlinespace
    {Edit Type:} & Paraphrase \\ \addlinespace
    {Question:} & Is the current conduction negatively affected by the amount of mobile charge carriers? \\ \addlinespace
    {Answer:} & Yes \\ \addlinespace
    \midrule
    {Extracted Edited Sentence:} & At the junction of a p-type and an n-type semiconductor there forms a depletion region where current conduction is inhibited by the absence of mobile charge carriers. \\
    \bottomrule
    \end{tabularx}
    \caption{
      \label{fig:condaqasample}
      An example from CondaQA.
      The negation in the original sentence is \emph{lack}.
      The crowdworkers wrote a paraphrase of the original sentence, which is included in the edited passage (\emph{[\ldots] by the absence of mobile charge carriers}).
      The question is written based on the original paragraph and answered based on the original and all three edited passages (only paraphrase edit shown).
      The answer to the question (for the edited passage) is \emph{Yes}.
      The dataset does not explicitly indicate the edited sentence.
      However, we extract it as explained in Appendix~\ref{sec:condaqaexample}.
    }
\end{figure*}

Figure~\ref{fig:condaqasample} shows an example from CondaQA.
Note that CondaQA highlights the original negated sentences from the original passages but not the edited sentences.
However, we use the available information in the dataset
such as the original sentence, the original passage and the edited passage
to extract the edited sentences.
Specifically, we identify sentence boundaries in the original passage
and pinpoint the index of the sentence that contains negation.
Then, we identify sentence boundaries in the edited passage
and use the same index to extract the edited sentence.
We use the extracted edited sentence to generate affirmative interpretations.
The authors manually analyzed 100 samples of the extracted edited sentences
and confirmed that in 96\% of the cases,
the extracted edited sentences are the same as the edited sentences in the passage.

\begin{table*}
  \small
  \centering
  \newcommand{\sig}{$^{\ast}$}

\small
\begin{tabular}{l r l l}
\toprule 
Edit        & \multicolumn{1}{c}{\% Negated} & \multicolumn{1}{c}{Meaning} & \multicolumn{1}{c}{Answer} \\
\midrule                                      
Paraphrase  & 59.5  & Same & Unchanged \\
Scope       & 97.7  & Changed  & Unchanged or changed \\
Affirmative & 43.6  & Reversed  & Reversed  \\
\bottomrule
\end{tabular}
  \caption{
    \label{table:condaqaedits} 
    Basic properties of the edits made by crowdworkers in the process of creating CondaQA.
    The \emph{Negated} column shows the percentage of edits that have negation.
    The \emph{Meaning} and \emph{Answer} columns indicate the differences in meaning (if any) between
    (1) the original and edited passage and
    (2) answers to the same question according to the original and edited passage.
    \emph{Changed} does not necessarily mean \emph{reversed}.
  }
\end{table*}

Additionally,
Table~\ref{table:condaqaedits} shows the basic properties of the edits made by crowdworkers.

\section{Additional Results with CondaQA}
\label{sec:condaqadetailed}

\begin{table*}
\small
  \centering
  \newcommand{\sig}{$^{\ast}$}

\setlength{\tabcolsep}{0.06in}
\small
\begin{tabular}{l l l l c@{\hspace{.15in}} r@{}l r@{}l r@{}l r@{}l@{\hspace{.06in}} r@{}l }
\toprule
&& \multicolumn{2}{c}{Input Representation} && \multicolumn{10}{c}{Accuracy}  \\
\cmidrule{3-4} \cmidrule{6-15} 
& \# Params. & \multicolumn{1}{c}{Training} & \multicolumn{1}{c}{Testing} && All && Ori. && Par. && Sco. && Aff. & \\
\midrule
~~~~RoBERTa-Large & 355M & Q & Q && 47.4 && 52.1 && 52.3 && 47.4 && 39.0 & \\
& & P & P && 45.4 && 46.5 && 46.1 && 45.2 && 43.9 & \\ \addlinespace
& & P+Q & P+Q && 64.9 && 67.2 && 66.0 && 59.5 && 66.0 & \\ 
~~~~~~~~w/ sentence with neg. from P (S) & & P+Q+S & P+Q+S && \cellcolor{green!2} 65.2 && \cellcolor{red!9} 66.0 && \cellcolor{red!11} 64.6 && \cellcolor{green!21} 61.8 && \cellcolor{green!19} 68.3 &\\ 
~~~~~~~~w/ 1st par. of S by \texttt{T5-CG}  (\texttt{S\textsubscript{CG}}) && P+Q+\texttt{S\textsubscript{CG}} & P+Q+\texttt{S\textsubscript{CG}} && \cellcolor{green!6} 65.7 && \cellcolor{green!9} 68.3 && \cellcolor{green!9} 67.1 && \cellcolor{green!6} 60.2 && \cellcolor{green!8} 67.0 &\\ 
\multicolumn{1}{l}{~~~~~~~~w/ Affirmative Interpretations} \\
& & P+Q+\texttt{A\textsubscript{HB}} & P+Q && \cellcolor{red!17} 62.8 && \cellcolor{red!21} 64.6 && \cellcolor{red!25} 62.9 && \cellcolor{red!8} 58.6 && \cellcolor{red!9} 64.9 & \\ 
& & P+Q+\texttt{A\textsubscript{HB}} & P+Q+\texttt{A\textsubscript{HB}} && \cellcolor{green!18} 67.1 & \sig & \cellcolor{green!10} 68.5 && \cellcolor{green!16} 68.0 && \cellcolor{green!21} 61.8 && \cellcolor{green!30} 69.7 & \sig \\
& & P+Q+\texttt{A\textsubscript{CG}} & P+Q && \cellcolor{red!30} 61.3 && \cellcolor{red!20} 64.7 && \cellcolor{red!30} 62.3 && \cellcolor{red!12} 58.2 && \cellcolor{red!51} 59.8 & \\
& & P+Q+\texttt{A\textsubscript{CG}} & P+Q+\texttt{A\textsubscript{CG}} && \cellcolor{green!12} 66.4 & \sig & \cellcolor{green!11} 68.6 && \cellcolor{green!10} 67.2 && \cellcolor{green!20} 61.7 && \cellcolor{green!14} 67.8 &\\
& & P+Q+\texttt{A\textsubscript{HB}}+\texttt{A\textsubscript{CG}} & P+Q+\texttt{A\textsubscript{HB}}+\texttt{A\textsubscript{CG}} && \cellcolor{green!5} 65.6 && \cellcolor{green!9} 68.4 && \cellcolor{green!4} 66.6 && \cellcolor{red!0} 59.4 && \cellcolor{green!13} 67.6 &\\ \addlinespace
& & P+Q+\texttt{A\textsubscript{G}} & P+Q && \cellcolor{red!11} 63.6 && \cellcolor{red!16} 65.2 && \cellcolor{red!10} 64.8 && \cellcolor{red!8} 58.6 && \cellcolor{red!4} 65.5 & \\
& & P+Q+\texttt{A\textsubscript{G}} & P+Q+\texttt{A\textsubscript{HB}} && \cellcolor{red!4} 64.4 && \cellcolor{red!13} 65.5 && \cellcolor{red!5} 65.3 && \cellcolor{green!7} 60.3 && \cellcolor{green!1} 66.2 & \\
& & P+Q+\texttt{A\textsubscript{G}} & P+Q+\texttt{A\textsubscript{CG}} && \cellcolor{green!5} 65.6 && \cellcolor{red!0} 67.2 && \cellcolor{green!6} 66.8 && \cellcolor{green!1} 59.7 && \cellcolor{green!18} 68.2 & \\ \addlinespace
& & P+Q+\texttt{A\textsubscript{G}} or \texttt{A\textsubscript{HB}} & P+Q && \cellcolor{red!20} 62.5 && \cellcolor{red!24} 64.2 && \cellcolor{red!21} 63.4 && \cellcolor{red!9} 58.5 && \cellcolor{red!19} 63.6 & \\
& & P+Q+\texttt{A\textsubscript{G}} or \texttt{A\textsubscript{HB}} & P+Q+\texttt{A\textsubscript{HB}} && \cellcolor{green!6} 65.7 && \cellcolor{red!0} \cellcolor{green!10} 67.2 && 67.2 && \cellcolor{green!0} 59.6 && \cellcolor{green!18} 68.2 & \\
& & P+Q+\texttt{A\textsubscript{G}} or \texttt{A\textsubscript{CG}} & P+Q && \cellcolor{red!36} 60.6 && \cellcolor{red!37} 62.6 && \cellcolor{red!35} 61.7 && \cellcolor{red!17} 57.6 && \cellcolor{red!47} 60.3 & \\
& & P+Q+\texttt{A\textsubscript{G}} or \texttt{A\textsubscript{CG}} & P+Q+\texttt{A\textsubscript{CG}} && \cellcolor{green!15} 66.7 & \sig & \cellcolor{green!14} 69.0 && \cellcolor{green!10} 67.2 && \cellcolor{green!26} 62.4 &\sig & \cellcolor{green!14} 67.8 & \\
\bottomrule
\end{tabular}
  \caption{
    \label{table:condaqadetailed} 
    The accuracy of RoBERTa-Large on the CondaQA test set for each edit type.
    We indicate statistically significant improvements (McNemar's test
    \cite{mcnemar1947note},
    $p < 0.05$) with respect to the model trained without affirmative interpretations
    (P+Q during training and testing) on each edit type with an asterisk (\sig).
  }
\end{table*}

Table~\ref{table:condaqadetailed} shows additional results 
with RoBERTa-Large and CondaQA for each edit type.
The results show that incorporating affirmative interpretations with RoBERTa-Large
improves results not only with the entire test set,
but also with each edit type individually.
However, not all of the improvements are statistically significant.
The only statistically significant improvements are
with (1) the scope edit type
when trained with P+Q+\texttt{A\textsubscript{CG}} or \texttt{A\textsubscript{CG}} and tested with P+Q+\texttt{A\textsubscript{CG}}, 
and
(2) the affirmative edit type
when trained with P+Q+\texttt{A\textsubscript{HB}} and tested with P+Q+\texttt{A\textsubscript{HB}}.

\section{UnifiedQA-v2 Training Corpora}
\label{sec:unifiedqacorpora}

\begin{table*}
\small
  \centering
  \setlength{\tabcolsep}{0.05in}
\small
\begin{tabular}{l r l}
\toprule 
\multicolumn{1}{l}{Corpus} & \multicolumn{1}{c}{\# Train Inst.} & \multicolumn{1}{l}{Reference} \\
\midrule
Squad 1.1 & 87,599 & \citet{rajpurkar-etal-2016-squad} \\
Squad 2 & 130,319 & \citet{rajpurkar-etal-2018-know} \\
Newsqa & 92,549 & \citet{trischler-etal-2017-newsqa} \\
Quoref & 19,399 & \citet{dasigi-etal-2019-quoref} \\
Ropes & 10,924 & \citet{lin-etal-2019-reasoning} \\
NarrativeQA & 32,747 & \citet{kocisky-etal-2018-narrativeqa} \\
DROP & 77,409 & \citet{dua-etal-2019-drop} \\
NaturalQuestions & 307,373 & \citet{kwiatkowski-etal-2019-natural} \\
MCTest & 1,480 & \citet{richardson-etal-2013-mctest} \\
RACE & 87,866 & \citet{lai-etal-2017-race} \\
OpenBookQA & 4,957 & \citet{mihaylov-etal-2018-suit} \\
ARC & 2,590 & \citet{clark2018think} \\
CommonsenseQA & 9,741 & \citet{talmor-etal-2019-commonsenseqa} \\
QASC & 8,134 & \citet{Khot_Clark_Guerquin_Jansen_Sabharwal_2020} \\
PhysicalIQA & 16,000 & \citet{bisk2019piqa} \\
SocialIQA & 33,410 & \citet{sap-etal-2019-social} \\
Winogrande & 40,398 & \citet{Sakaguchi_LeBras_Bhagavatula_Choi_2020} \\
BoolQ & 9,427 & \citet{clark-etal-2019-boolq} \\
MultiRC (yes/no) & 6,000 & \citet{khashabi-etal-2018-looking}\\
BoolQ-NP & 9,727 & \citet{khashabi-etal-2020-bang} \\
\bottomrule
\end{tabular}
  \caption{
    \label{table:unifiedqacorpora} 
    The corpora that \citet{khashabi2022unifiedqav2} used to train UnifiedQA-v2,
    and the number of training instances in each corpus.
  }
\end{table*}

Table~\ref{table:unifiedqacorpora} shows the QA corpora 
that \citet{khashabi2022unifiedqav2} used to train UnifiedQA-v2.
These corpora span the following QA formats:
extractive, abstractive, multiple-choice, and yes-no questions.

\end{document}